%% file: temporal_prediction.tex
\documentclass[sigconf, nonacm=true]{acmart}

\usepackage{multirow}
\usepackage{hyperref}

\usepackage[utf8]{inputenc} 

\AtBeginDocument{%
  \providecommand\BibTeX{{%
    \normalfont B\kern-0.5em{\scshape i\kern-0.25em b}\kern-0.8em\TeX}}}

\setcopyright{acmcopyright}
\copyrightyear{2020}
\acmYear{2020}
\acmDOI{000000000}

\acmConference[KDD '20]{ACM SIGKDD International Conference on Knowledge Discovery \& Data Mining}{August 22--27, 2020}{San Diego, CA}
\acmBooktitle{ACM SIGKDD International Conference on Knowledge Discovery \& Data Mining, August 22--27, 2020, San Diego, CA}
\acmPrice{15.00}
\acmISBN{978-1-4503-XXXX-X/18/06}

\begin{document}

\title{PlumeNet: Large-Scale Air Quality Forecasting Using A Convolutional LSTM Network}







\author{Antoine Alléon}
\affiliation{\institution{Plume Labs}}
\email{antoine.alleon@plumelabs.com}

\author{Grégoire Jauvion}
\affiliation{\institution{Plume Labs}}
\email{gregoire.jauvion@plumelabs.com}

\author{Boris Quennehen}
\affiliation{\institution{Plume Labs}}
\email{boris.quennehen@plumelabs.com}

\author{David Lissmyr}
\affiliation{\institution{Plume Labs}}
\email{david.lissmyr@plumelabs.com}


\begin{abstract}

This paper presents an engine able to forecast jointly the concentrations of the main pollutants harming people's health: nitrogen dioxyde (NO$_{2}$), ozone (O$_{3}$) and particulate matter (PM$_{2.5}$ and PM$_{10}$, which are respectively the particles whose diameters are below $2.5~\mu m$ and $10~\mu m$ respectively).

The forecasts are performed on a regular grid (the results presented in the paper are produced with a $0.5^{\circ}$ resolution grid over Europe and the United States) with a neural network whose architecture includes convolutional LSTM blocks. The engine is fed with the most recent air quality monitoring stations measures available, weather forecasts as well as air quality physical and chemical model (AQPCM) outputs. The engine can be used to produce air quality forecasts with long time horizons, and the experiments presented in this paper show that the $4$ days forecasts beat very significantly simple benchmarks.

A valuable advantage of the engine is that it does not need much computing power: the forecasts can be built in a few minutes on a standard GPU. Thus, they can be updated very frequently, as soon as new air quality measures are available (generally every hour), which is not the case of AQPCMs traditionally used for air quality forecasting.

The engine described in this paper relies on the same principles as a prediction engine deployed and used by Plume Labs in several products aiming at providing air quality data to individuals and businesses.

\end{abstract}




\maketitle

\input{content_external}


\bibliographystyle{ACM-Reference-Format}
\bibliography{temporal_prediction}










\end{document}

%% file: content_external.tex
\section{Introduction}



Air pollution is one of the major public health concern. World Health Organization (WHO) estimates that more than 80\% of citizens living in urban environments where air quality is monitored are exposed to air quality levels that exceed WHO guideline limits. It also estimates that $4.2$ million deaths every year are linked to outdoor air pollution \cite{WHO_report} exposure.

Despite those alarming figures, very few citizens have access to information about the quality of the air they breathe. More and more public and private initiatives are being developed to close this gap and give to citizens the information they need to protect themselves from air pollution.

This is a particularly challenging topic because air quality varies a lot, both in time and in space. For example, a polluted air can become clean in a few hours after a heavy rain. Also, a crowded street can be much more polluted than a green park area a few hundreds meters away \cite{AQ_review}.

This paper focuses on temporal air quality forecasts and presents an engine able to forecast the concentrations of atmospheric pollutants regulated by WHO: nitrogen dioxide (NO$_{2}$), ozone (O$_{3}$) and particulate matter (PM$_{2.5}$ and PM$_{10}$, which are respectively the particles whose diameters are below $2.5~\mu m$ and $10~\mu m$ respectively). This is a key topic as having acces to accurate air quality estimates ahead of time enables individuals to plan outdoor activities accordingly.

Traditionally, temporal air quality forecasts are based on air quality physical and chemical models (AQPCMs hereafter): they rely on physical and chemical modeling of pollutants' emissions, chemical reactions and dispersion, and the simulations are initiated with monitoring stations and satellite-based measurements. They often rely on human-made assumptions and are able to model accurately specific events impacting air quality, like wildfires or large-scale pollution transport. Their main drawback is that they need huge amounts of computing power and they are complex to set up.

Our forecasting engine is based on a neural network architecture which builds forecasts on a regular grid: the results presented in the paper use a $0.5^{\circ}$ (approximately $50$ kilometers) resolution grid over Europe and the United States. We use an encoder-decoder architecture based on several convolutional LSTM blocks which are known to perform well for long-term spatiotemporal predictions. It is fed with air quality monitoring stations measurements collected in the last hours, weather forecasts as well as AQPCM outputs.

The processes which drive the evolution of air pollutant concentrations within the atmosphere, i.e., the transport, removal and chemical reactivity of pollutants, are significantly influenced by meteorology \cite{atmospheric}. Thus, weather forecasts are commonly used to forecast air quality. The AQPCM outputs have shown a significant impact on forecasts' accuracy, in particular because of their ability to forecast specific and extreme air pollution events.

The paper is organized as follows. We discuss earlier works in Section 2. Section 3 gives a detailed overview of the data sources used by the engine. Section 4 presents the architecture of the forecasting engine and details the model estimation process. Section 5 provides an evaluation of the forecasting engine.

\section{Related work}

The problem of air quality prediction is much studied in the literature and is tackled through various angles. \cite{bigdata_summary} and \cite{deep_learning_review} present comprehensive reviews of air quality modeling using machine learning approaches.

\cite{deepplume}, \cite{global_approach} and \cite{r_package_airpred} focus on spatial modeling and do not take into account air pollution temporal variability. They model the main air pollutants spatial variability at a given time using diverse datasets including monitoring stations measurements, satellite-based measurements, land-use datasets and traffic datasets. They reach very fine resolutions, from 10 meters in \cite{deepplume} to a few kilometers in \cite{global_approach} and \cite{r_package_airpred}.

Other papers focus on air quality temporal variability and aim at predicting air pollutants future concentrations at a given location. A commonly used framework consists in building air quality forecasts at a given monitoring station using the station's past measurements as well as weather forecasts at the station's location. The models are learnt on historical datasets, and most of recent papers using this approach are based on neural networks, and more specifically on LSTM architectures. \cite{deepair} builds air quality forecasts in Beijing up to 10 hours using an encoder-decoder LSTM architecture. \cite{deep_air_net} and \cite{india_lstm_rnn} apply slightly different recurrent architectures in a few Indian cities.

More similarly to the approach we use, several papers use spatiotemporal modeling frameworks to take into account both spatial and temporal variability. \cite{airnet} introduces a dataset with air quality and meteorological data on a $0.25^{\circ}$ regular grid in China over $2$ years.
\cite{conv_lstm_2}, \cite{deep_air_learning}, \cite{deep_air_quality}, \cite{fine-grained}, \cite{spatial_temporal} and \cite{deep_distributed} build air quality forecasts at China's monitoring stations using the stations' past measurements and weather forecasts. Spatial correlations between the stations are included in a deep learning architecture, and the temporal variability is modeled with LSTM layers in \cite{conv_lstm_2}, \cite{deep_air_learning} and \cite{spatial_temporal}. \cite{u-air} uses a similar approach but integrates land-use and traffic features in the forecasts.
In \cite{conv_lstm}, the authors build air quality forecasts on a regular grid in the Beijing area and in the London area using a convolutional LSTM architecture similar to the one we use in this paper. In this paper as well, air quality forecasts are based on past measurements and weather forecasts. \cite{att_conv_lstm} uses an architecture with both convolutional LSTM and attention mechanisms.
In order to limit overfitting due to the low number of monitoring stations available, other papers like \cite{physical_model} and \cite{model_data_driven} introduce physical modeling of the pollutants dispersion in the spatiotemporal model.

In \cite{precipitation_nowcasting} and \cite{metnet}, the authors build precipitation forecasts with a very fine resolution ($1$ kilometer) over the United States using very similar spatiotemporal architectures: \cite{precipitation_nowcasting} uses a U-net network and \cite{metnet} uses attention mechanisms. The convolutional LSTM block is introduced in \cite{conv_lstm_theory} and applied on precipitation forecasting.

The encoder-decoder LSTM architecture we use in this paper is also very commonly used in language translation (\cite{seq_learning}).

\section{Data sources}

This section details the data sources used by the forecasting engine.

\subsection{Air quality measurements}

\begin{figure}[h]
  \centering
  \includegraphics[width=\linewidth]{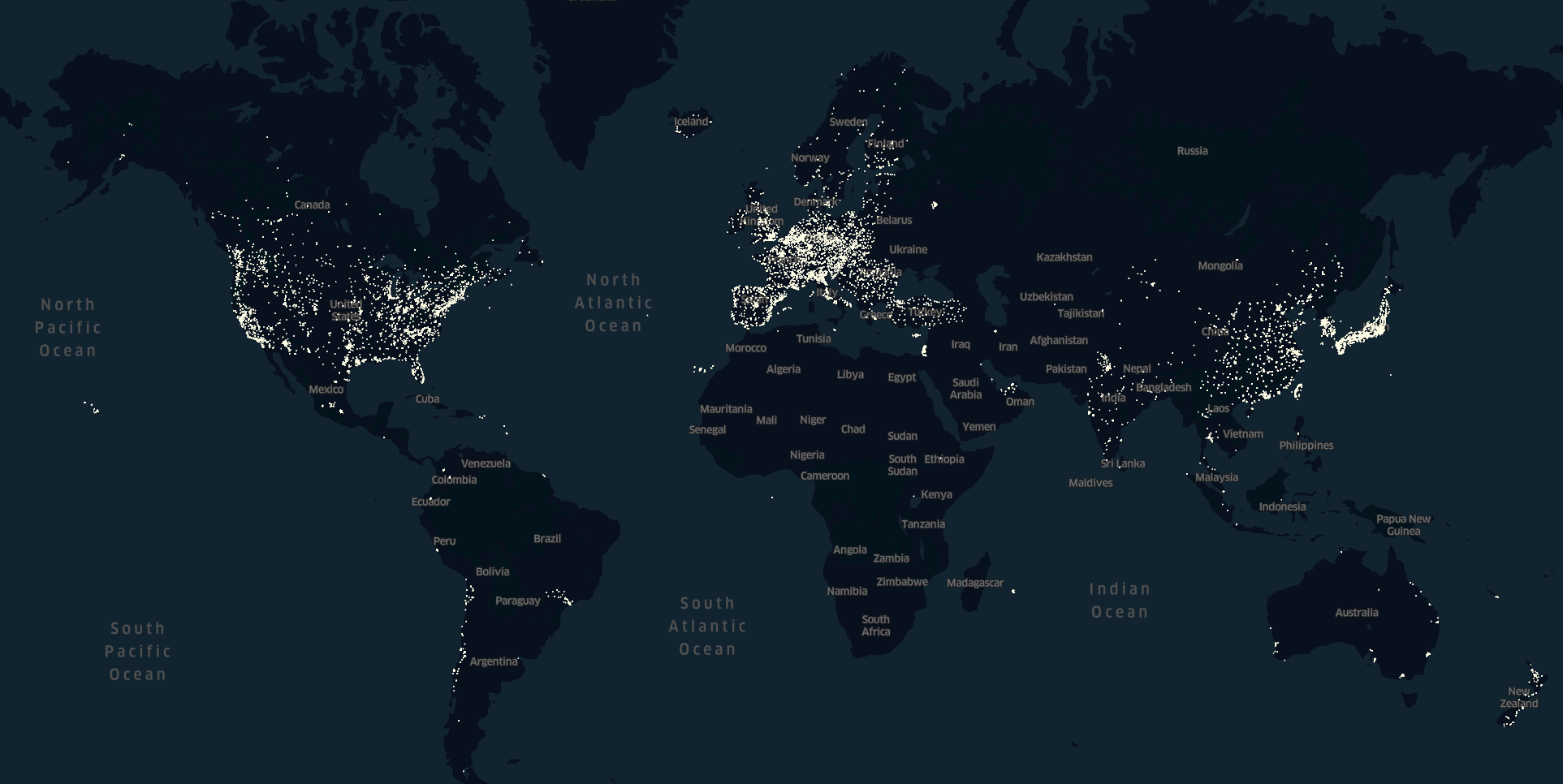}
  \caption{Map of the monitoring stations whose measurements are included in the predictions}
  \label{fig:stations}
  \Description{This map shows the locations of the monitoring stations whose air quality measurements feed the prediction engine.}
\end{figure}

We have built a proprietary architecture based on several dozens of crawlers collecting the air quality measurements provided by about $14000$ monitoring stations across the world. Figure~\ref{fig:stations} shows a global map of the locations of those monitoring stations.

Almost all monitoring stations measurements are available on a hourly basis. Each monitoring station does not necessarily measure the four pollutants predicted by the engine. The experiments presented in this paper have been performed in Europe and the United States, which are the $2$ regions where the number of measurements we have collected is the highest. Table~\ref{tab:stations} gives the number of monitoring stations as well as the number of stations measuring each pollutant in those regions.

\begin{table}
  \caption{Number of monitoring stations per region and pollutant}
  \label{tab:stations}
  \begin{tabular}{c||c|cccc}
    \toprule
    Region & Global & NO2 & O3 & PM2.5 & PM10 \\
    \midrule
    Europe & 2778 & 2252 & 1614 & 790 & 1958 \\
    United States & 1924 & 331 & 1300 & 1041 & 347 \\
  \bottomrule
\end{tabular}
\end{table}

We include in the datasets all measurements from January 1st, 2019 to December 31st, 2019. We have found that there are missing and erroneous values (generally abnormally high) coming from those monitoring stations. They can be encountered during station maintenance windows, during station failures or if issues arise during the publishing or collection of said data. While we are not able to determine the exact cause of such errors, it is important to detect them and define an appropriate treatment: missing values are discarded from the datasets, and erroneous values are detected using an outlier detection engine and then discarded.

\subsection{Weather forecasts}

Weather simulations are computed using physical and chemical modeling of natural (atmospheric and land-soil) phenomenon. The weather forecasts provide numerous features like temperature, wind, precipitation, soil moisture or snow depth. In the experiments presented in this paper, we have used the following features: temperature, relative humidity, wind speed and direction (encoded in $u$ and $v$ which are respectively the wind zonal and meridional velocities), planetary boundary layer height and precipitation rate.

The weather forecasts used here are produced daily on a regular grid covering the whole world with a surface resolution of $28$ kilometers. The time horizon is $16$ days, with a degressive time resolution of $1$ hour for the first $5$ days, $3$ hours between $6$ and $10$ days and $12$ hours above. We have collected them once a day from January 1st, 2019 to December 31st, 2019, and the first timestep of the forecasts is midnight in UTC time.

\subsection{AQPCM outputs}

AQPCMs rely on physical and chemical modeling of pollutants' emissions, chemical reactions and dispersion, and the simulations are initiated with monitoring stations and satellite-based measurements. AQPCMs are characterised by their geographical coverage (regional or global) and their spatial granularity (from a few to dozens of kilometres).

The AQPCM forecasts used in this paper have a $0.4^{\circ}$ spatial resolution and a $3$ hours time resolution on a $5$ days horizon. They cover the $4$ pollutants forecasted by the engine (NO$_{2}$, O$_{3}$, PM$_{2.5}$ and PM$_{10}$). We have collected them once a day from January 1st, 2019 to December 31st, 2019, and the first timestep of the forecasts is midnight UTC time.

\section{Description of the forecasting engine}

The engine builds air quality forecasts on a regular grid built over the region of interest. We note $(N_x, N_y)$ the grid shape. The data sources used (air quality measurements, weather forecasts and AQPCM outputs) are projected onto this grid. We note $N_{pol}=4$ the number of pollutants forecasted by the engine, which is also the number of outputs of the AQPCM ($1$ output per pollutant), and $N_{features}$ the number of weather features used.

In the experiments presented in this paper, the forecasts are built on a $3$ hours basis: this choice limits the size and complexity of the datasets and hence the training and inference times for long forecasting horizons. It is worth noting that the engine can also be used on a hourly basis which is the frequency of most of air quality monitoring stations measurements.

We note $N_{in}$ the number of timesteps of historical air quality measurements fed in the engine, and $N_{out}$ the number of timesteps forecasted, which is also the number of timesteps of weather forecasts and AQPCM outputs fed to the engine. As an example, $N_{in}=8$ and $N_{out}=32$ means that the engine is fed with the last $24$ hours of air quality measurements, $96$ hours weather forecasts and AQPCM outputs, and produces $96$ hours air quality forecasts.

\subsection{Datasets' description}

We introduce the euclidean distance $\| l - l' \|$ between two locations $l$ and $l'$. We define also the exponential kernel $k_d(l, l') = exp(-\frac{\| l - l' \|}{d})$, where the distance $d$ is expressed in kilometers.

Every $3$ hours from January 1st, 2019 to December 31st, 2019, an observation is produced with the following data:
\begin{itemize}
  \item Historical air quality measurements in the last $N_{in}$ timesteps projected on the engine's grid. At a given time, each grid cell is a weighted average of the available measurements around, and the weights are computed with the exponential kernel $k_d$. We use $d = 100$ kilometers, which is high enough to make sure that the weights are significantly greater than $0$ everywhere on the grid, and low enough to not smooth too much the spatial variability of the pollutants concentrations
  \item The last produced weather forecasts and AQPCM outputs in the next $N_{out}$ timesteps. Those forecasts are produced on a different grid than the grid used by the engine, and they are projected on the engine's regular grid using a bilinear interpolation
\end{itemize}

\subsection{Architecture of the engine}

Convolutional LSTM (ConvLSTM hereafter) is a powerful block to model spatiotemporal data with strong correlations in space. A ConvLSTM determines the future state of a given grid cell by using the inputs and past states of its local neighbors: this is achieved by using a convolution operator in the state-to-state and input-to-state transitions. This enables to reduce very significantly the number of parameters of the block compared to a fully-connected LSTM block where every grid cell would be connected to all inputs and past states on the whole grid. An important feature of the ConvLSTM block is that its number of parameters does not depend on the size of the spatial grid but only on the number of hidden states and on the size of the convolution kernels.

\begin{figure*}[h]
  \centering
  \includegraphics[width=\linewidth]{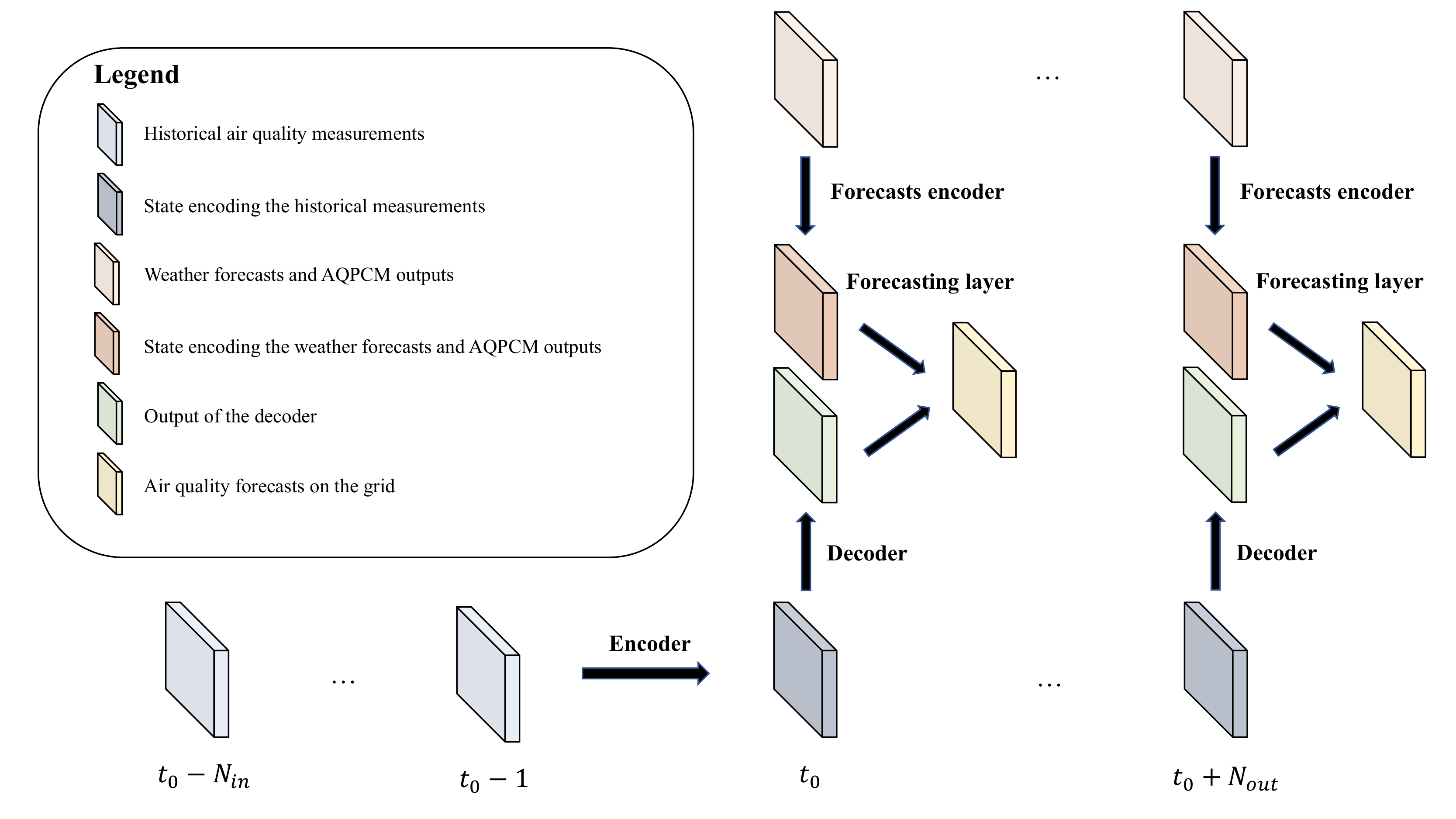}
  \caption{Neural network architecture}
  \label{fig:architecture}
  \Description{Neural network architecture.}
\end{figure*}

We use ConvLSTM as a building block of our architecture shown in Figure~\ref{fig:architecture}, which is made up of four parts:
\begin{itemize}
  \item The encoder, the forecasts encoder and the decoder are formed with one or several successive ConvLSTM blocks. We note respectively $H_{enc}$, $H_{f\_enc}$ and $H_{dec}$ the number of hidden states in the last ConvLSTM block of the encoder, the forecasts encoder and the decoder

  \begin{itemize}
    \item The encoder inputs the sequence of historical air quality measurements projected on the grid of shape $(N_{in}, N_x, N_y, N_{pol})$ and outputs a state encoding the historical data of shape $(N_x, N_y, H_{enc})$
    \item The forecasts encoder inputs the sequence of weather forecasts and AQPCM outputs (projected on the grid) of shape $(N_{out}, N_x, N_y, N_{features} + N_{pol})$, and outputs a sequence of states, of shape $(N_{out}, N_x, N_y, H_{f\_enc})$. At every timestep, the state encodes the forecasts until this timestep
    \item The decoder inputs a constant sequence equal to the encoder output, of shape $(N_{out}, N_x, N_y, H_{enc})$ and outputs a sequence of states, of shape $(N_{out}, N_x, N_y, H_{dec})$. Using a constant sequence as input in the decoder is a classical choice in encoder-decoder architectures
  \end{itemize}

  \item At every of the $N_{out}$ forecast timesteps, the forecasting block inputs a concatenation of the states produced by the decoder and the forecasts encoder, of shape $(N_x, N_y, H_{dec} + H_{f\_enc})$, and outputs the air pollutants concentrations, of shape $(N_x, N_y, N_{pol})$. This is done by using a simple $2$-dimensional convolutional layer with $N_{pol}$ $(1,1)$ kernels and a relu activation function
\end{itemize}

A batch normalization layer is used after each ConvLSTM block.

\subsection{Model estimation}

The datasets are splitted into an evaluation dataset formed with all observations in March, June, September and December and a training dataset formed with the remaining months. We also ensure that the time ranges used to evaluate the model have not been used at all during training by removing observations in the training set which have less than $4$ days of difference with the closest observation in the evaluation set. For example, having day $D$ in training set and day $D+1$ in the evaluation set is problematic as the $4$ days forecasts performed at those dates have $3$ days in common.

Also, using an evaluation set formed with a continuous time period may have biased the results because of the high seasonality of air pollution. That is why the evaluation set is formed with $4$ different months regularly spaced in time.

Every observation in the training and evaluation sets is a tensor of size $(N_{in}, N_x, N_y, N_{features} + N_{pol})$. Those inputs as well as the intermediate tensors needed to train the model have a large memory footprint. Thus, the model is trained on smaller subgrids of size $(20,20)$. The drawback of this approach is that it prevents from modeling pollution transport on a very large scale greater than the size of this subgrid. However, we noticed that it worked much better in practice.

The encoder, the forecasts encoder as well as the decoder have the same architecture and are formed with $2$ ConvLSTM blocks with $64$ and $32$ units respectively. We use $(3,3)$ convolution kernels.

The model has been implemented with Keras and Tensorflow and trained on a Nvidia Geforce GTX $1080$. The loss used in training is the mean squared logarithmic error (MSLE) loss. We use Adam optimizer with a learning rate equal to $0.001$. The batch size is $16$. The number of epochs is $20$.

\section{Evaluation of the forecasting engine}

This section provides a detailed analysis of the accuracy of the forecasts produced by the engine in Europe and the United States, with the parameters given in the previous section. The evaluation metric is the MSLE loss function used to train the models. The engine is compared to $2$ benchmark models:
\begin{itemize}
  \item A simple benchmark, \textit{Constant benchmark}, which consists in assuming that on every point of the grid the pollutants concentrations remain constant in the future, equal to the last know value. This may seem to be an overly simplistic benchmark to compare to, however it is common practice when working on air quality or weather forecasting given the difficulty of those tasks
  \item An other similar benchmark, \textit{Constant benchmark adjusted}, where the constant forecasts are adjusted at every hour of the day with a multiplicative factor modeling the daily seasonality of air pollution. At each timestep, the factor is equal to the ratio between the average concentrations at this hour of the day and the hour when the forecasts are computed (i.e. corresponding to the last known value). The adjustment factors are computed per pollutant and grid cell. This benchmark performs significanlty better than the naive \textit{Constant benchmark}
\end{itemize}

We considered also benchmarks based on AQPCM outputs but they performed actually worse than the two benchmarks introduced above: the main reason is that the forecasts produced by AQPCMs are not built to forecast the weighted average of monitoring stations measurements as we do in our approach. Thus, they are not comparable to our engine's forecasts.

\subsection{Forecasts accuracy}

\begin{figure}[h]
  \centering
  \includegraphics[width=\linewidth]{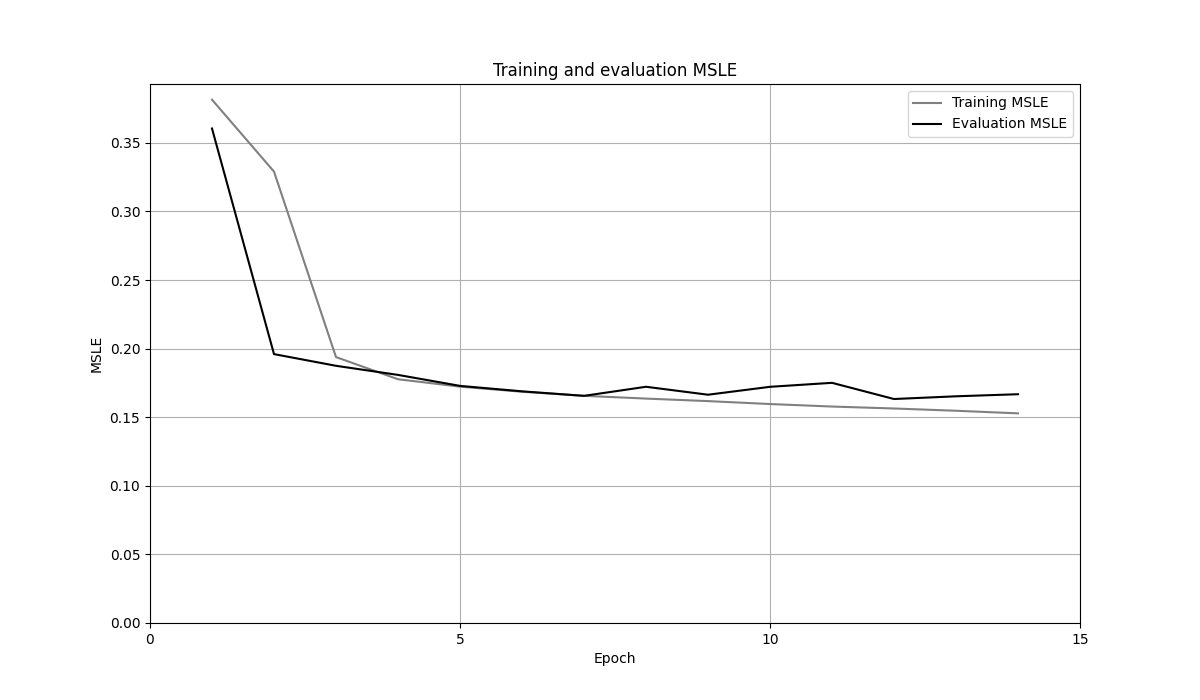}
  \caption{Loss as a function of the number of epochs}
  \label{fig:loss}
  \Description{TODO}
\end{figure}

\begin{figure}[h]
  \centering
  \includegraphics[width=\linewidth]{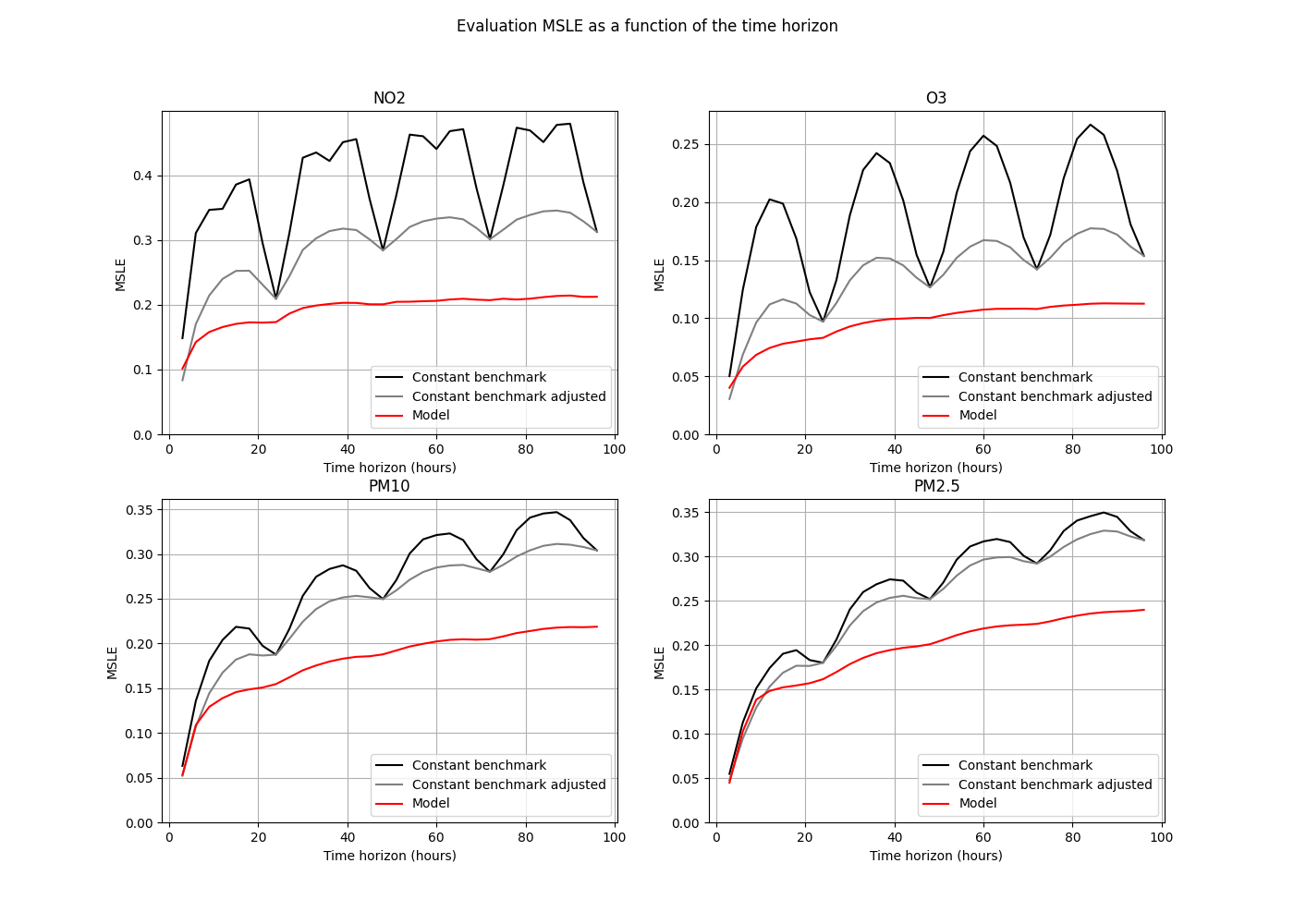}
  \caption{Loss as a function of the time horizon}
  \label{fig:time_horizon_loss}
  \Description{TODO}
\end{figure}

Figure~\ref{fig:loss} shows the train and evaluation loss as a function of the number of epochs, and Figure~\ref{fig:time_horizon_loss} shows the evaluation loss as a function of the time horizon for each pollutant (in Europe). Tables~\ref{tab:accuracy} and ~\ref{tab:accuracy_us} give the evaluation loss averaged until 24 hours and 96 hours in Europe and the United States respectively. We see that the prediction model gives a very significant improvement compared to the benchmarks for all pollutants, in particular for long time horizons.

\begin{table}
  \caption{Loss averaged over the time horizon (Europe)}
  \label{tab:accuracy}
  \begin{tabular}{c|ccc|ccc}
    \hline
     & \multicolumn{3}{c}{24 hours} & \multicolumn{3}{c}{96 hours} \\
     & Cst & Cst adj. & Model & Cst & Cst adj. & Model \\
    \hline
    NO2 & 0.318 & 0.206 & 0.146 & 0.389 & 0.289 & 0.193 \\
    O3 & 0.149 & 0.091 & 0.063 & 0.189 & 0.138 & 0.096 \\
    PM10 & 0.174 & 0.147 & 0.119 & 0.266 & 0.244 & 0.181 \\
    PM2.5 & 0.151 & 0.135 & 0.116 & 0.259 & 0.247 & 0.194 \\
    \hline
    Global & 0.198 & 0.145 & 0.111 & 0.276 & 0.230 & 0.166
\end{tabular}
\end{table}

\begin{table}
  \caption{Loss averaged over the time horizon (United States)}
  \label{tab:accuracy_us}
  \begin{tabular}{c|ccc|ccc}
    \hline
     & \multicolumn{3}{c}{24 hours} & \multicolumn{3}{c}{96 hours} \\
     & Cst & Cst adj. & Model & Cst & Cst adj. & Model \\
    \hline
    NO2 & 0.269 & 0.195 & 0.154 & 0.374 & 0.304 & 0.194 \\
    O3 & 0.174 & 0.079 & 0.060 & 0.206 & 0.110 & 0.076 \\
    PM10 & 0.245 & 0.215 & 0.175 & 0.356 & 0.332 & 0.223 \\
    PM2.5 & 0.117 & 0.099 & 0.088 & 0.202 & 0.189 & 0.126 \\
    \hline
    Global & 0.201 & 0.147 & 0.119 & 0.284 & 0.234 & 0.155
\end{tabular}
\end{table}

\subsection{Impact of the features fed to the engine}

In this section, we compare the following models to quantify how each of the input features impacts the forecasts accuracy:
\begin{itemize}
  \item Model using $24$ hours of historical air quality measurements, weather forecasts and AQPCM outputs (\textit{All features})
  \item Model using the last known historical air quality measurements only, weather forecasts and AQPCM outputs (\textit{Without histo. measurements})
  \item Model using $24$ hours of historical air quality measurements and weather forecasts (\textit{Without AQPCM outputs})
  \item Model using $24$ hours of historical air quality measurements and AQPCM outputs (\textit{Without weather forecasts})
\end{itemize}

Table~\ref{tab:features} gives the evaluation loss averaged until $24$ hours of those different models. We see that all the inputs fed to the engine (historical air quality measurement, weather forecasts and AQPCM outputs) have a very significant impact on the forecasts accuracy.

\begin{table*}
  \caption{Average loss until $24$ hours for models using different features (Europe)}
  \label{tab:features}
  \begin{tabular}{c|cccc}
    \hline
     & All features & Without histo. measurements & Without AQPCM outputs & Without weather forecasts \\
    \hline
    NO2 & 0.146 & 0.169 & 0.155 & 0.168 \\
    O3 & 0.063 & 0.076 & 0.069 & 0.081 \\
    PM10 & 0.119 & 0.128 & 0.129 & 0.132 \\
    PM2.5 & 0.116 & 0.125 & 0.136 & 0.137 \\
    \hline
    Global & 0.111 & 0.125 & 0.122 & 0.130 \\
    \hline

\end{tabular}
\end{table*}

\subsection{Impact of the size of the convolution kernels}

All ConvLSTM blocks in the engine's architecture use $(3,3)$ kernels. Table~\ref{tab:kernels} shows the evaluation loss averaged until $96$ hours of the models using $(1,1)$ and $(5,5)$ kernels.

The models using $(3,3)$ and $(5,5)$ kernels perform significantly better than the one using $(1,1)$ kernels: this means that the architecture is effective in capturing spatiotemporal patterns in pollutants concentrations, which can not be done with $(1,1)$ kernels.
The model using $(3,3)$ kernels performs better than the one using $(5,5)$ kernels: it comes probably from the much higher complexity and number of parameters of models using large convolution kernels.

\begin{table}
  \caption{Average loss until $96$ hours for models using different kenel sizes (Europe)}
  \label{tab:kernels}
  \begin{tabular}{c|ccc}
    \hline
     & (1,1) kernels & (3,3) kernels & (5,5) kernels \\
    \hline
    NO2 & 0.222 & 0.193 & 0.197 \\
    O3 & 0.113 & 0.096 & 0.104 \\
    PM10 & 0.197 & 0.181 & 0.189 \\
    PM2.5 & 0.209 & 0.194 & 0.206 \\
    \hline
    Global & 0.185 & 0.166 & 0.174 \\
    \hline
\end{tabular}
\end{table}

\subsection{Illustrations}

\begin{figure*}[h]
  \centering

  \includegraphics[width=.19\linewidth]{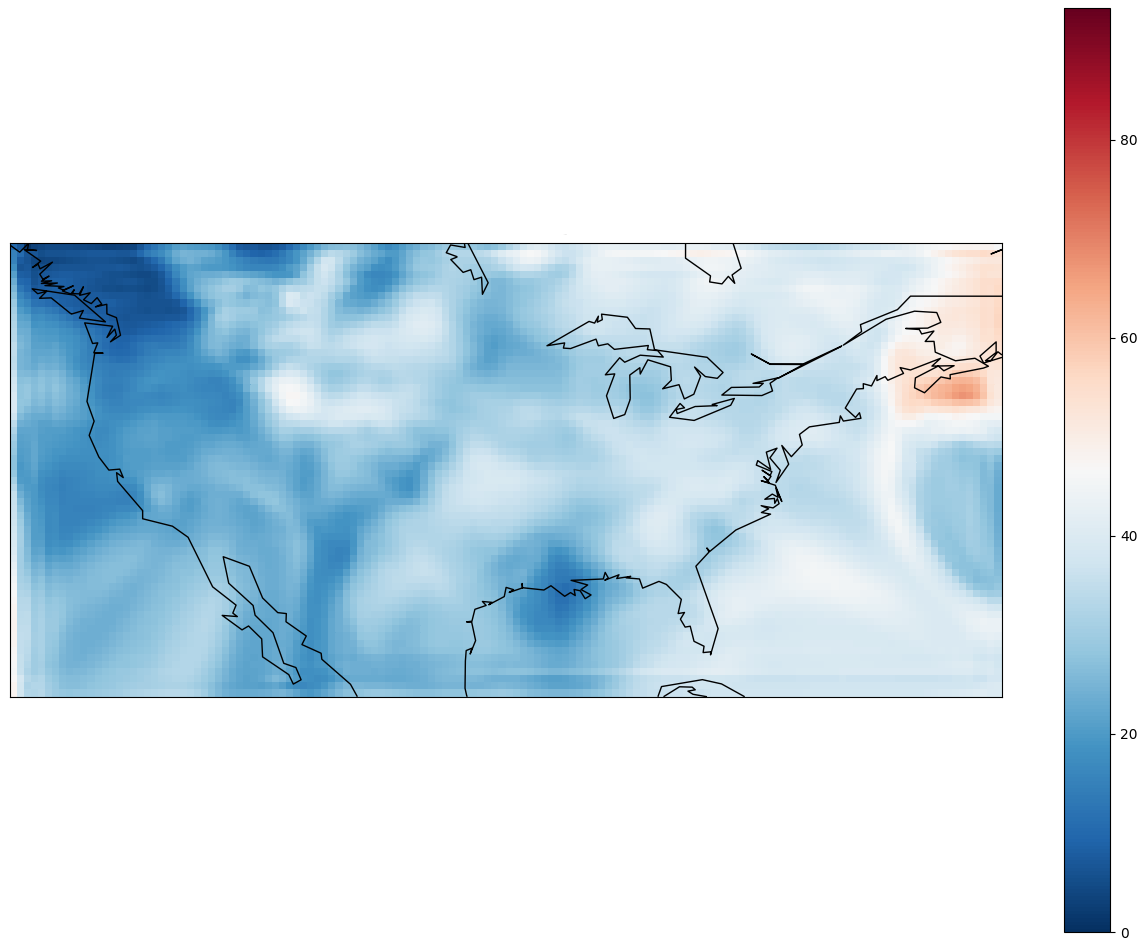}
  \includegraphics[width=.19\linewidth]{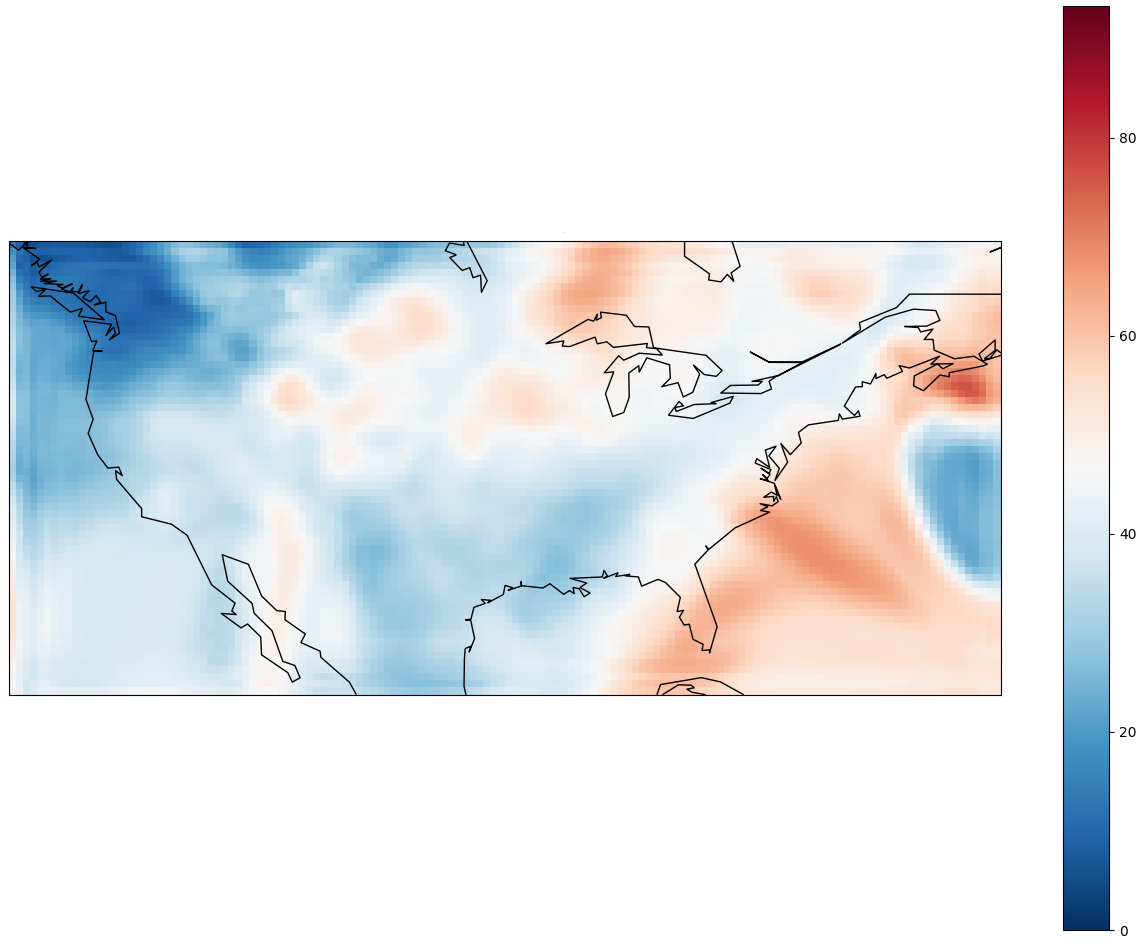}
  \includegraphics[width=.19\linewidth]{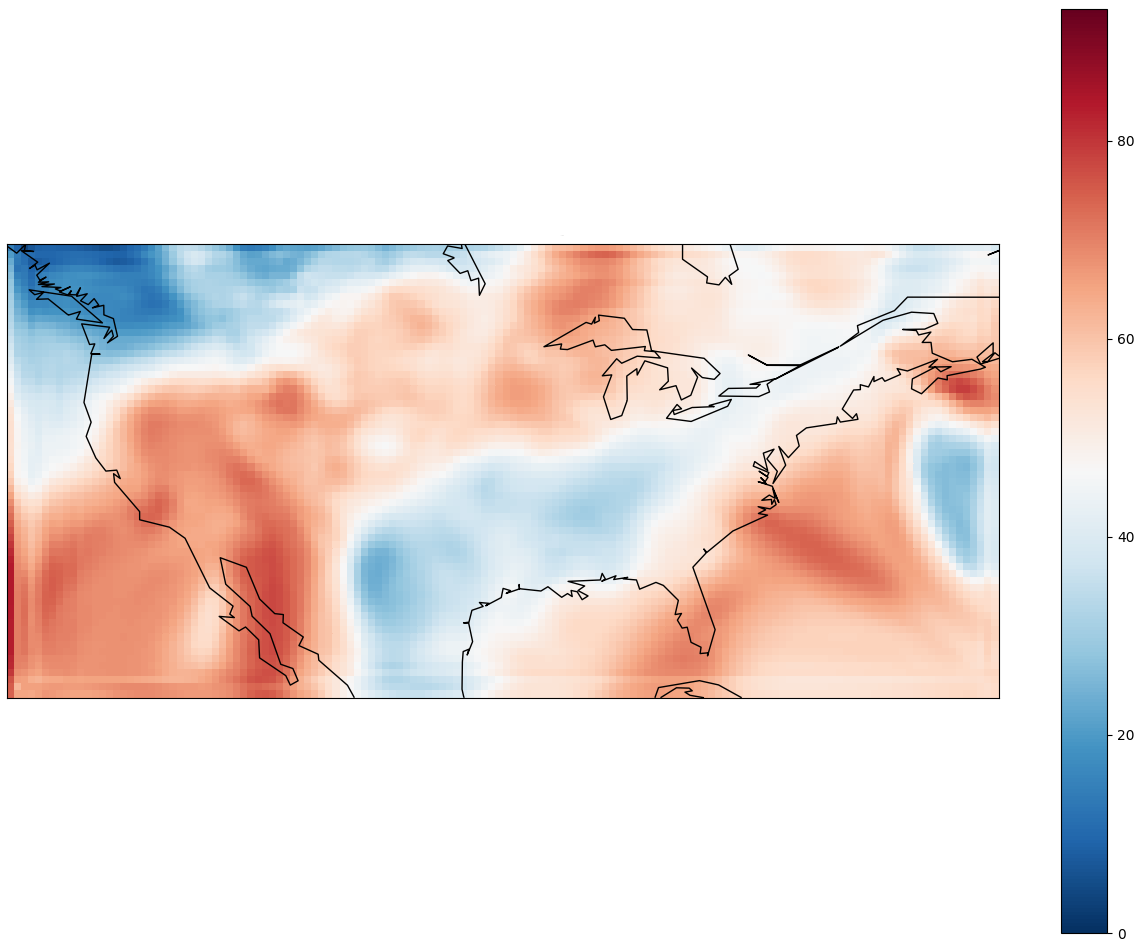}
  \includegraphics[width=.19\linewidth]{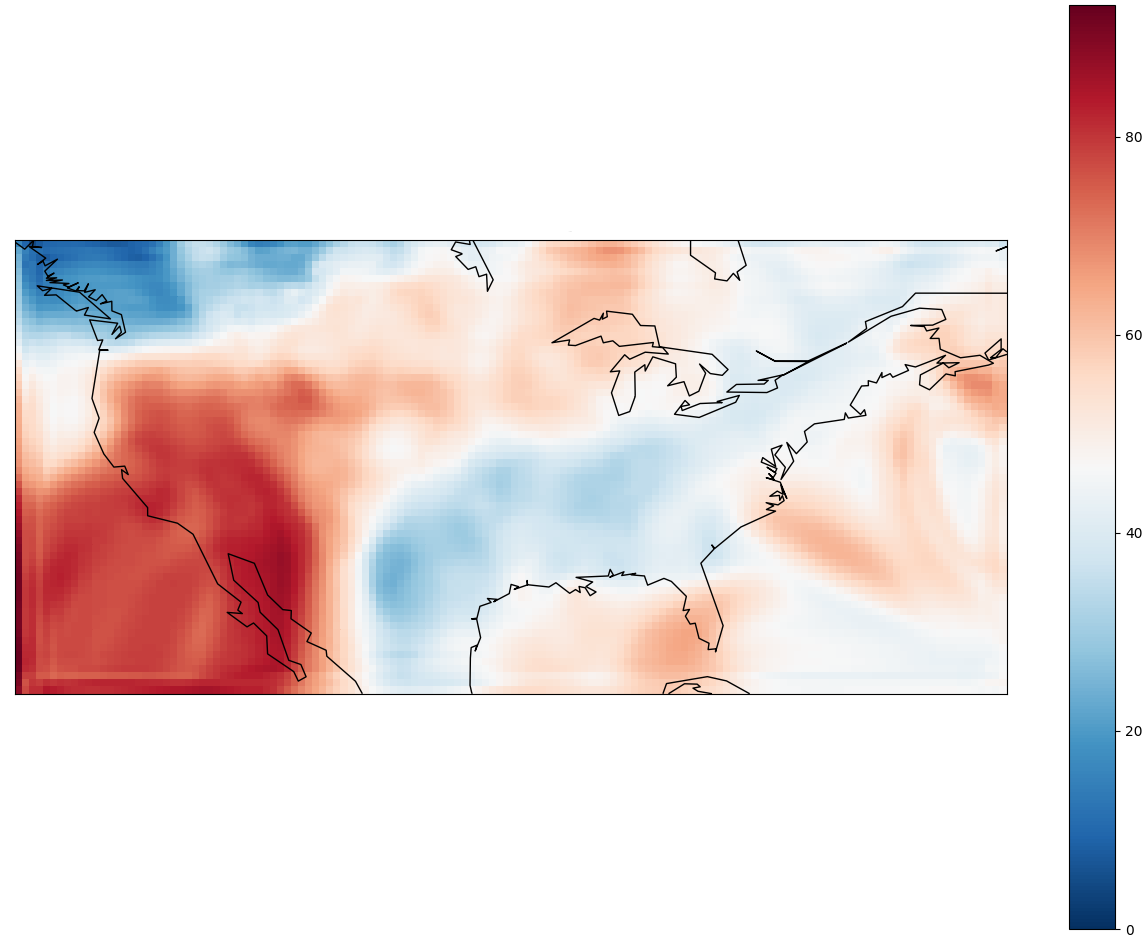}
  \includegraphics[width=.19\linewidth]{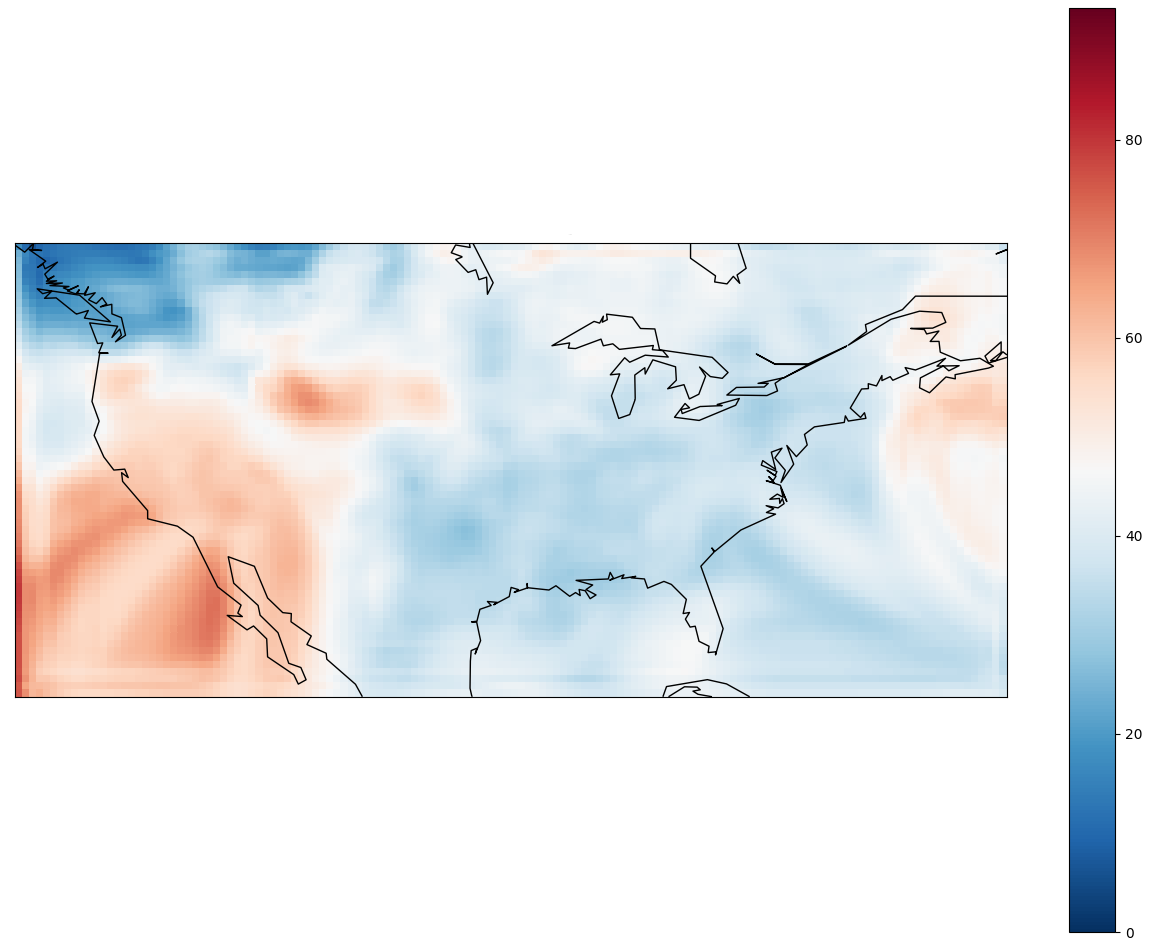}

  \caption{O$_3$ forecasts at $12$, $15$, $18$, $21$ and $24$ hours horizons (United States, 11/8/2019)}
  \label{fig:o3_forecasts}
  \Description{O$_3$ forecasts at $12$, $15$, $18$, $21$ and $24$ hours horizons (United States, 11/8/2019)}
\end{figure*}

\begin{figure*}[h]
  \centering

  \includegraphics[width=.19\linewidth]{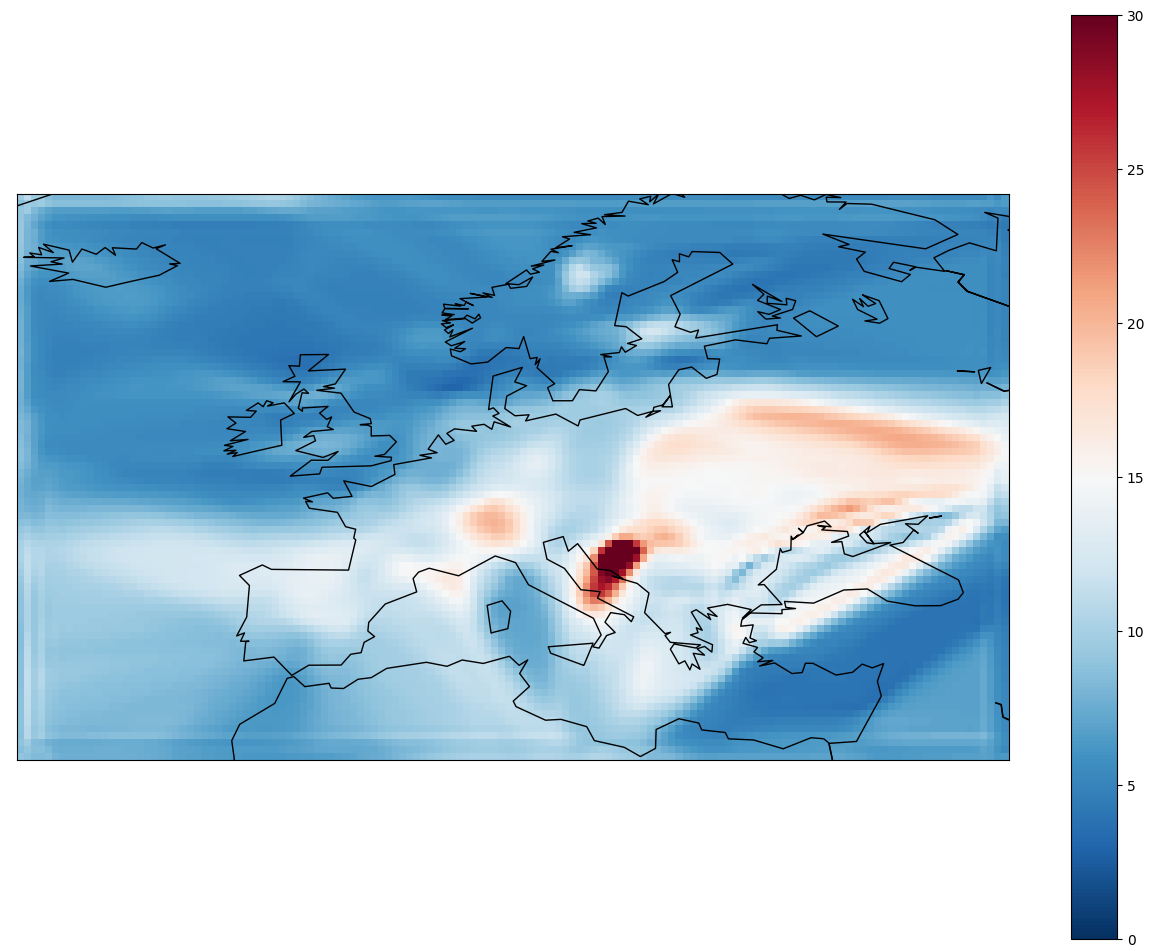}
  \includegraphics[width=.19\linewidth]{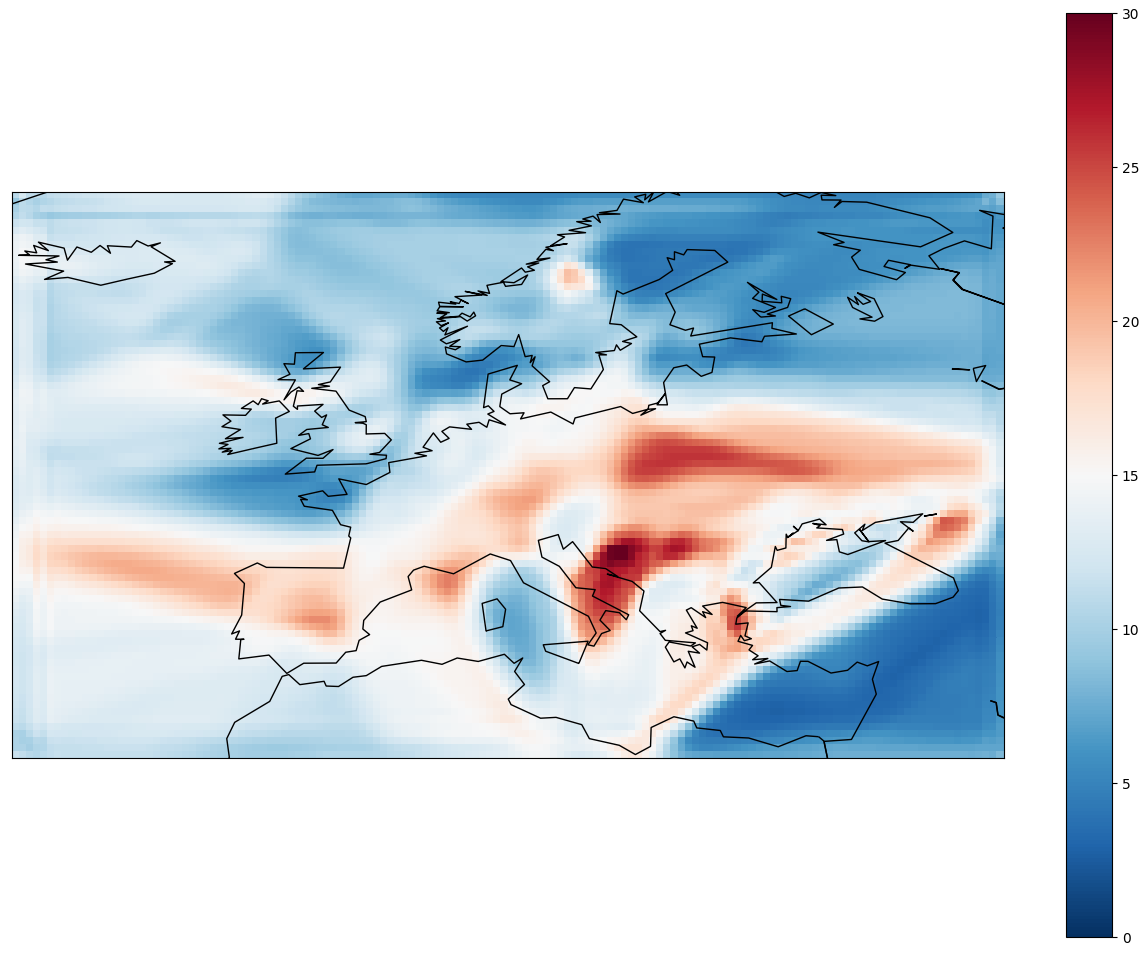}
  \includegraphics[width=.19\linewidth]{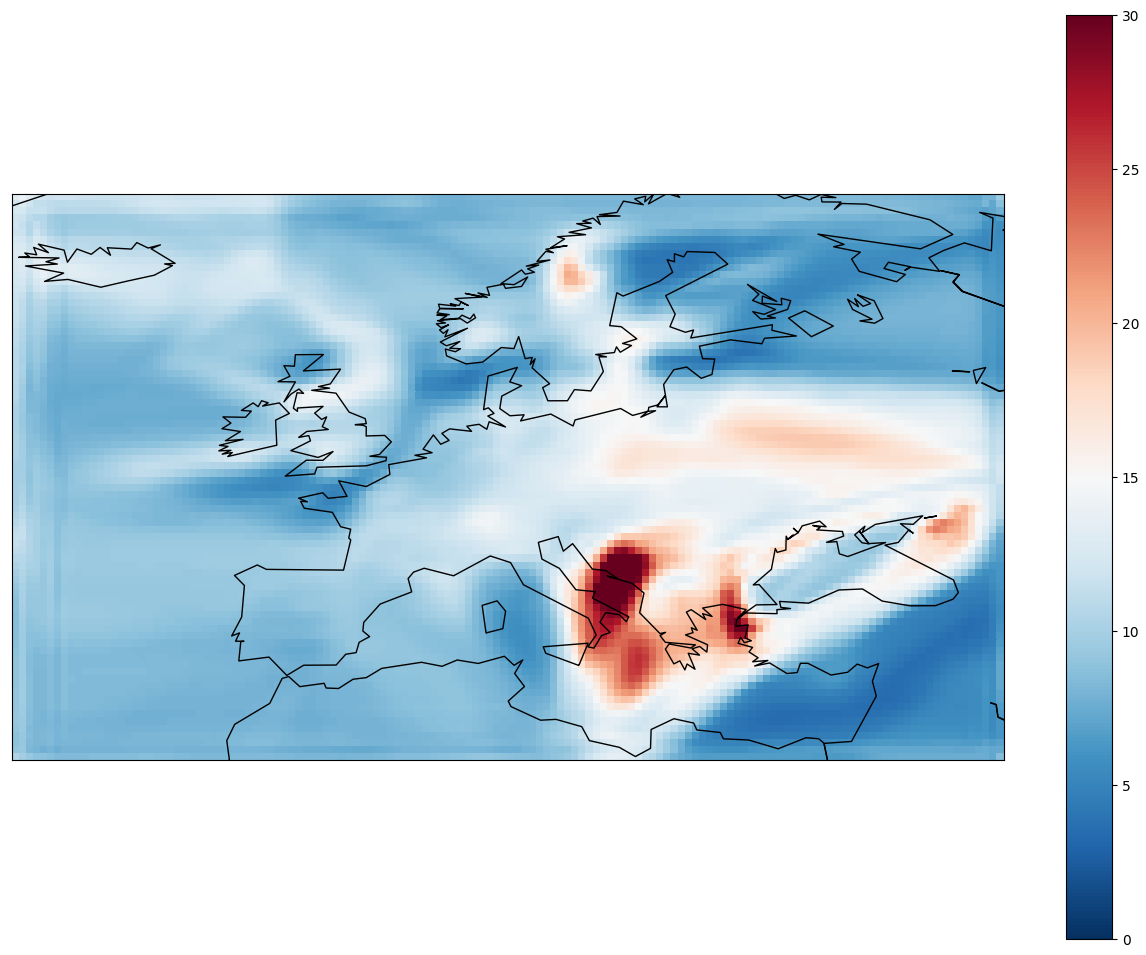}
  \includegraphics[width=.19\linewidth]{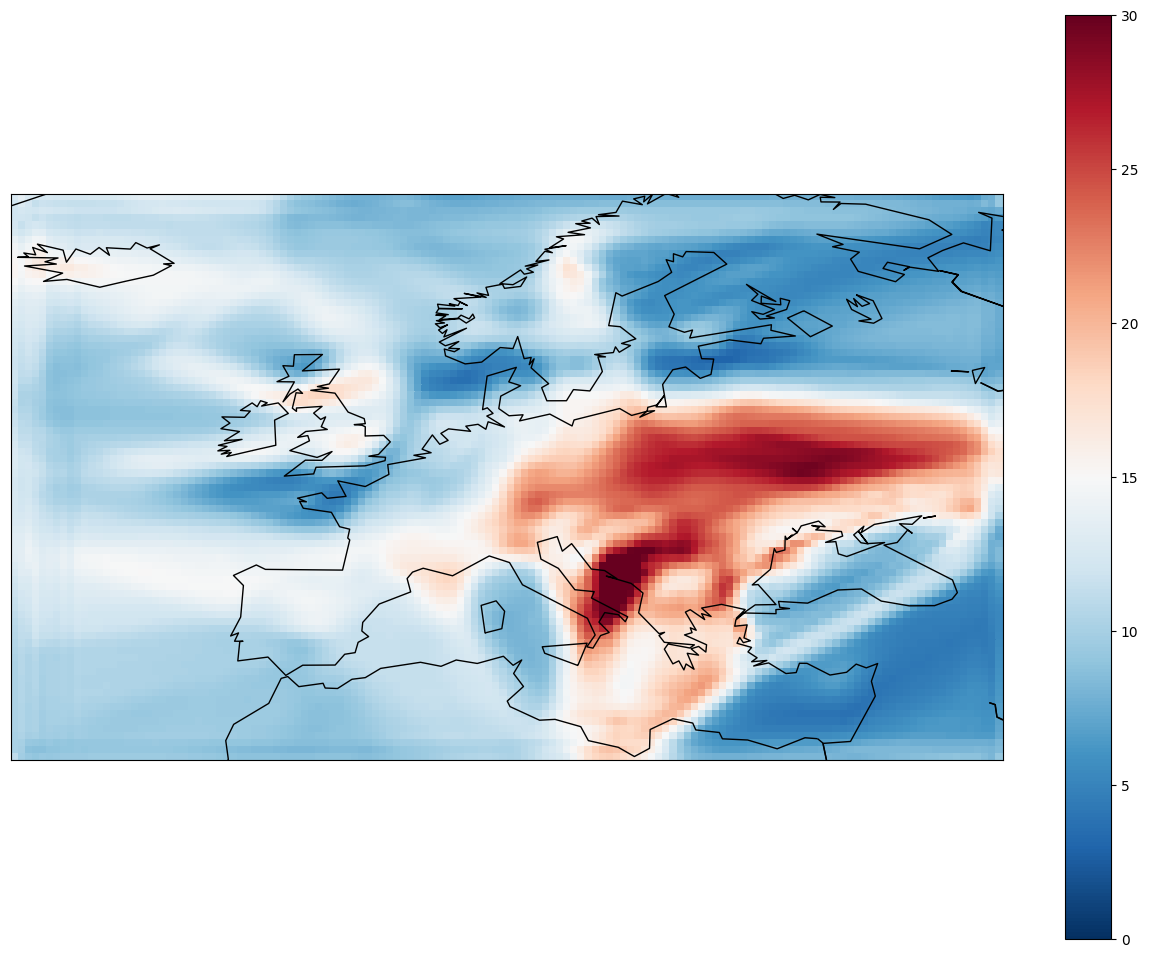}
  \includegraphics[width=.19\linewidth]{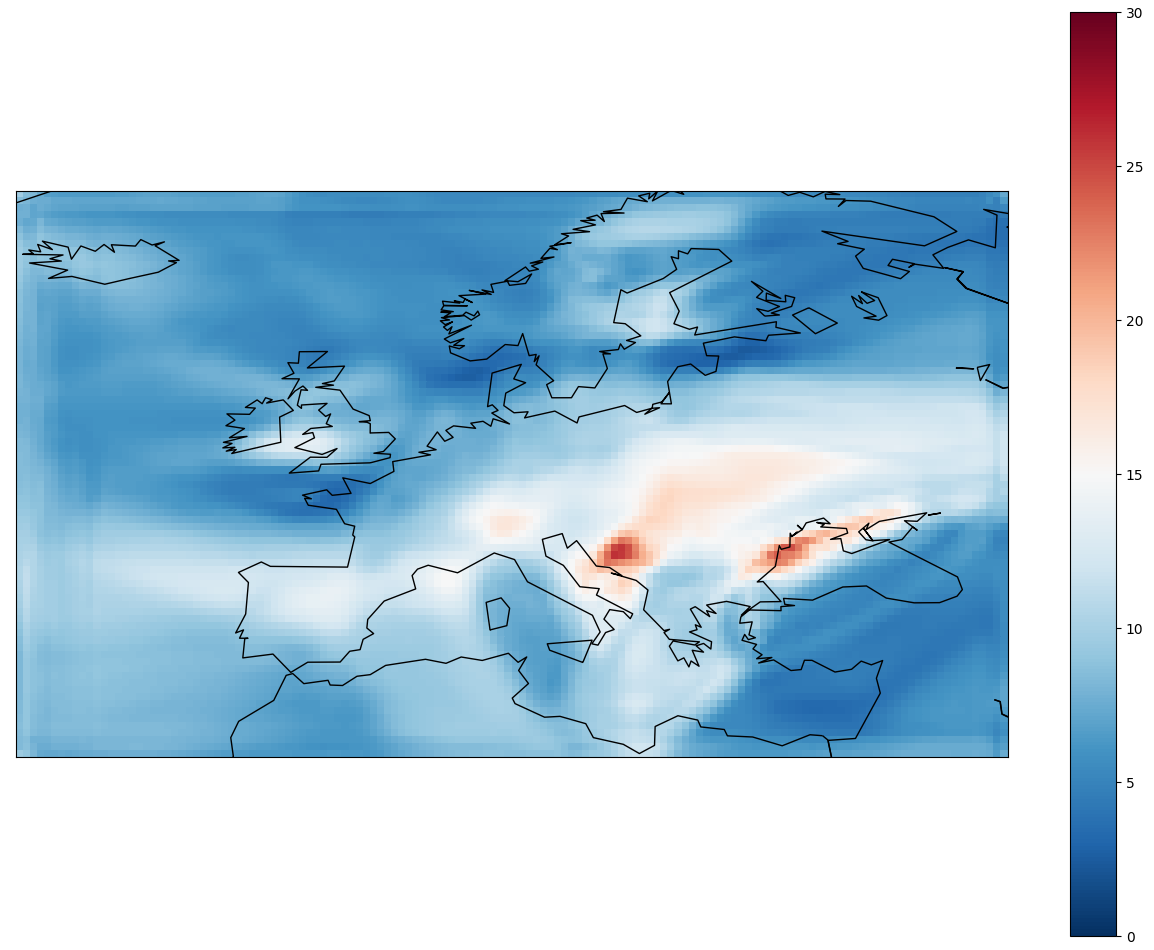}

  \caption{NO$_2$ forecasts at $0$, $6$, $12$, $18$ and $24$ hours horizons (Europe, 3/29/2020)}
  \label{fig:no2_forecasts}
  \Description{NO$_2$ forecasts at $0$, $6$, $12$, $18$ and $24$ hours horizons (Europe, 3/29/2020)}
\end{figure*}

As an illustration, Figures~\ref{fig:o3_forecasts} and ~\ref{fig:no2_forecasts} show O$_{3}$ forecasts produced in the United States on 11/8/2019 and NO$_2$ forecasts produced in Europe on 3/29/2020 respectively.

\section{Conclusion and future work}

The engine presented in this paper is at our knowledge the first air quality forecasting engine integrating at the same time air quality measurements, weather forecasts and AQPCM outputs. As detailed in the paper, each one of those inputs brings a lot of predictive power. The use of weather forecasts is somewhat classical to build air quality forecasts given the strong links between weather and air pollution. However, the use of AQPCM outputs is more innovative and enables to make the most of state-of-the-art atmospheric science modeling based on very strong human expertise.

Also, most existing works focusing on temporal air quality forecasts using machine learning approaches cover limited surfaces like Beijing or London metropolitan areas. Here, air quality forecasts are built over grids covering larger areas, and the approach can be replicated on a global grid, at the cost of an increased computing power needed to train the model and infer the predictions.

A very valuable advantage of the engine compared to AQPCM is that it needs much less computing power. As an illustration, producing $4$ days forecasts in Europe or the United States at a given time takes a few minutes with a standard GPU. This enables to update the forecasts very frequently while AQPCM outputs are generally produced one or two times a day. Hence, the engine's forecasts can integrate the most recent air quality measurements which is a very important feature.

An other useful feature of the recurrent architecture used is that the model can be used for long time horizons, as long as weather forecasts and AQPCM outputs are available. Weather forecasts are generally available at large time horizons. On the contrary, AQPCM outputs are generally limited to a few days, but they can be extrapolated to the wanted horizon using simple assumptions, at the cost of a lower accuracy of the engine.

We think that the engine can still be improved by using larger training datasets. Weather forecasts, AQPCM outputs and air quality measurements are all hard to gather at old dates, and that is why our datasets were limited to the year 2019, but they will get bigger as time goes by. In particular, larger datasets would enable the use of deeper and larger architectures than the one described in this paper, which would probably improve the forecasts accuracy at the cost of a higher need of computing power.

Finally, this engine focuses on large-scale air pollution modeling, and uses a coarse spatial resolution of $0.5^{\circ}$. This enables to cover large areas and is adapted to the density of the air quality monitoring network. We know that air quality can vary singificantly in a few dozens of meters. Adding additional and more granular features like traffic data, land-use features or power plants emissions would enable to model the spatial variability on a finer scale. It would also need much more computing power and make the model less tractable.